# VAEs and GANs: Implicitly Approximating Complex Distributions with Simple Base Distributions and Deep Neural Networks— Principles, Necessity, and Limitations

Yuan-Hao Wei
HKPU
Yuan-Hao.Wei@outlook.com

## Abstract

This tutorial focuses on the fundamental architectures of Variational Autoencoders (VAE) and Generative Adversarial Networks (GAN), disregarding their numerous variations, to highlight their core principles. Both VAE and GAN utilize simple distributions, such as Gaussians, as a basis and leverage the powerful nonlinear transformation capabilities of neural networks to approximate arbitrarily complex distributions. The theoretical basis lies in that a linear combination of multiple Gaussians can almost approximate any probability distribution, while neural networks enable further refinement through nonlinear transformations. Both methods approximate complex data distributions implicitly. This implicit approximation is crucial because directly modeling high-dimensional distributions explicitly is often intractable. However, the choice of a simple latent prior, while computationally convenient, introduces limitations. In VAEs, the fixed Gaussian prior forces the posterior distribution to align with it, potentially leading to loss of information and reduced expressiveness. This restriction affects both the interpretability of the model and the quality of generated samples.

## 1. Introduction

Generative models aim to learn the underlying data distribution and generate new, realistic samples. Among them, Variational Autoencoders (VAEs) (Kingma and Welling (2013)) and Generative Adversarial Networks (GANs) (Goodfellow et al. (2020)) have emerged as two of the most influential approaches. Despite their differences in architecture and training objectives, both models share a fundamental mechanism: they approximate complex data distributions implicitly, using a simple latent distribution (e.g., a standard normal distribution) combined with a deep neural network transformation.

This implicit approximation is crucial because directly modeling high-dimensional distributions explicitly is often intractable. For example, in image generation, the probability distribution of all possible facial images is too complex to be mathematically formulated. Instead, VAEs and GANs sample from a simple base distribution and apply a neural network to transform these samples into realistic outputs.

However, the choice of a simple latent prior, while computationally convenient, introduces limitations. In VAEs, the fixed Gaussian prior forces the posterior distribution to align with it, potentially leading to loss of information and reduced expressiveness in the latent space. This restriction affects both the interpretability of the model and the quality of generated samples, often resulting in blurry outputs. Similarly, in GANs, the reliance on a simple latent distribution may impact the model's ability to capture complex variations in the data.

This paper explores the principles, necessity, and limitations of implicit approximation in VAEs and GANs. We first analyze their shared mechanism of using simple priors and neural network mappings. Then, we discuss why implicit approximation is necessary for high-dimensional generative modeling. Finally, we examine the trade-offs involved in using simple priors, highlighting how they impact both inference and generation. By understanding these limitations, we can explore potential improvements, such as adaptive latent priors, to enhance both the interpretability and quality of generative models.

## 2. Why Are VAE and GAN So Similar?

The overall architecture and objective function of VAE are illustrated in Figure 1. VAE generates samples approximating the target dataset $\mathbf{X}_{data}$ by first drawing latent variables from the posterior distribution $q_\phi(z|x)$ and then passing them through a decoder (Doersch (2016)). In the standard VAE (vanilla VAE), the prior distribution $p(z)$ of the latent variables is set to a standard normal distribution. Due to the KL divergence term in the objective function, the posterior $q_\phi(z|x)$ is forced to be as close as possible to the prior distribution, i.e., a standard normal distribution. Therefore, the VAE generation process can be summarized as a combination of **a posterior distribution that is constrained to approximate a normal distribution and a neural network (decoder) that provides a nonlinear mapping**, ultimately producing samples equivalent to those in the target dataset $\mathbf{X}_{data}$.

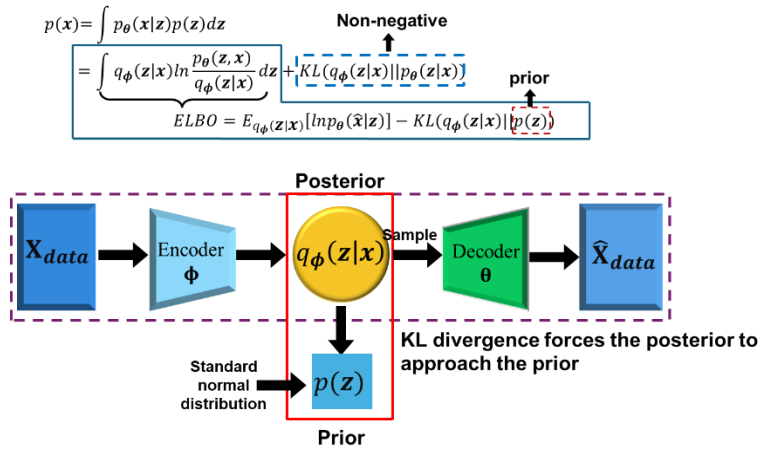

Figure 1. Illustration of VAE.

The overall architecture of Generative Adversarial Networks (GAN) is shown in Figure 2. GAN employs a discriminator to distinguish between samples from two distributions: one consisting of real samples directly drawn from the target dataset $\mathbf{X}_{data}$, and the other generated by first sampling from a standard normal distribution and then passing the samples through a generator. When the discriminator can no longer differentiate between these two distributions, it indicates that the combination of **a normal distribution and a neural network (generator) can implicitly approximate the distribution of** $\mathbf{X}_{data}$.

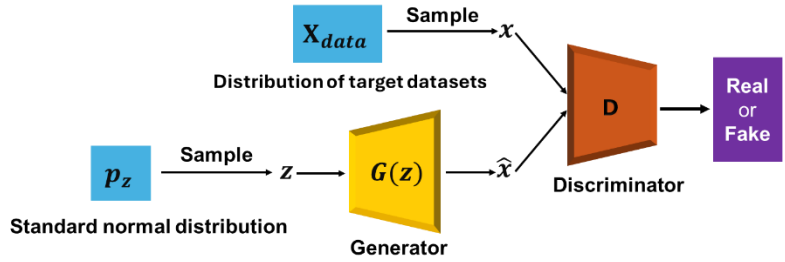

Figure 2. Illustration of GAN.

Figure 3 compares the generation processes of VAE and GAN. In VAE, the generation (decoding) process involves sampling from a posterior distribution that is forced to approximate a standard normal distribution, followed by a neural network (decoder) mapping. In contrast, GAN directly samples from a standard normal distribution and then applies a neural network (generator) transformation. Essentially, **both models utilize a simple basis distribution combined with a neural network to perform a complex transformation of the sampled points, thereby generating samples approaching the target dataset $X_{data}$**. In this framework, the simple distribution provides the fundamental components, while the neural network determines how these components are assembled, ultimately approximating the target distribution.

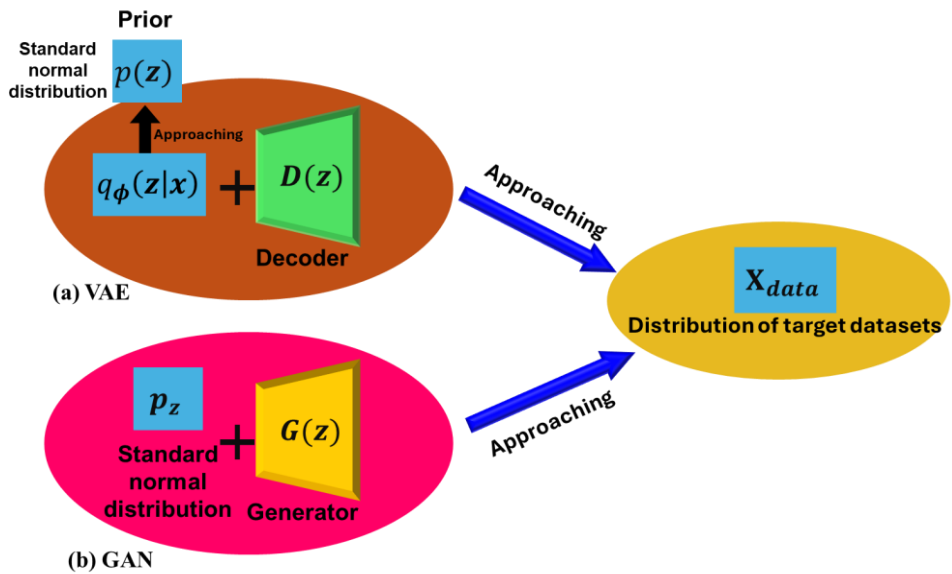

Figure 3. Generation process of VAE and GAN.

Figure 4 provides an analogy for the distribution approximation process in VAE and GAN. The simple basis distribution can be thought of as basic building blocks, while the neural network serves as the assembly instructions, and the target distribution represents the final complex structure formed by assembling the blocks. The complex structure (i.e., the distribution of $X_{data}$) is obtained by combining simple building blocks (the basis distribution) with structured assembly rules (the generator/decoder).

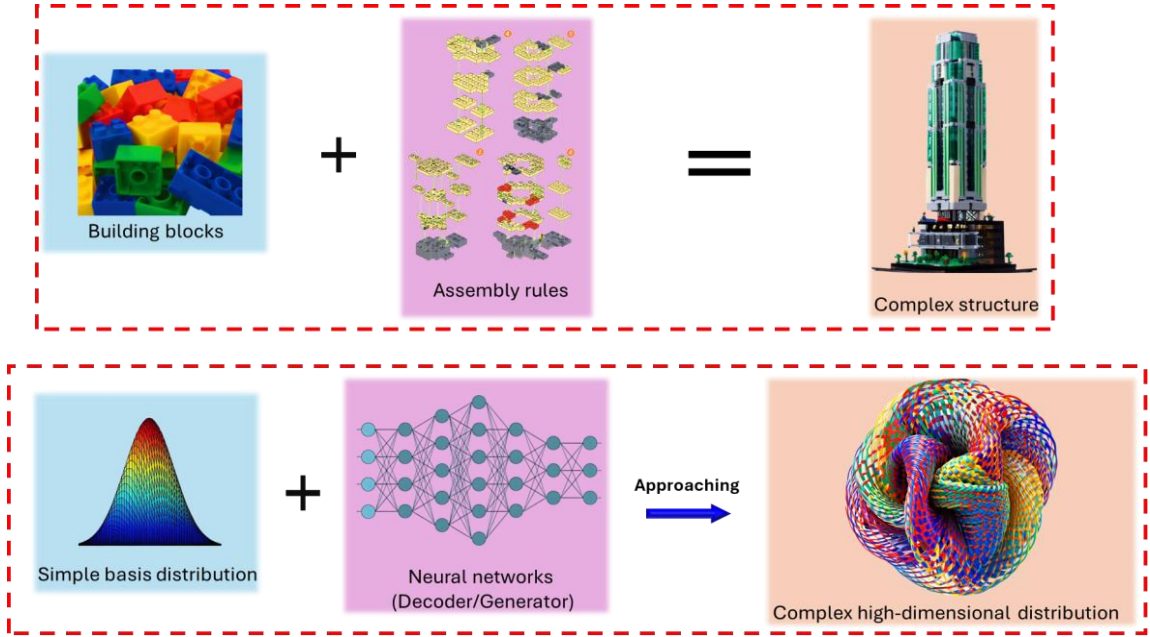

Figure 4. The complex structure (i.e., complex high-dimensional distribution) is obtained by combining simple building blocks (the basis distribution) with structured assembly rules (the generator/decoder).

## 3. The Necessity of Implicit Approximation in Generative Models

Neither VAE nor GAN have the capability to explicitly express the distribution of a target dataset $\mathbf{X}_{data}$. Instead, they approximate this distribution **implicitly** by generating samples that resemble those in $\mathbf{X}_{data}$. This implicit approximation is not just a limitation but a necessity, as many high-dimensional and complex distributions cannot be explicitly formulated.

For example, consider the distribution of facial images. Even for a single individual, there exists an infinite number of possible images due to variations in head orientation, facial expressions, lighting conditions, and aging effects. **Capturing all possible images of a single person using an explicit probability distribution is already infeasible, let alone modeling the distribution of all human faces explicitly.**

Therefore, the implicit approximation approach adopted by VAE and GAN is both **practical and effective**. By leveraging simple base distributions (e.g., Gaussian) and neural networks, these models can learn to generate data that statistically aligns with the target distribution without needing an explicit mathematical representation. This method enables generative models to handle high-dimensional and complex data distributions in a scalable and computationally feasible manner.

## 4. Advantages and Limitations of Using Simple Distributions (e.g., Normal Distribution) as the Latent Prior in Generative Models

Since the goal of VAE and GAN is to generate samples that resemble the target dataset $\mathbf{X}_{data}$, sampling is an essential step in their generation process. Both models first sample from a basis distribution (latent variable distribution) and then pass the sampled values through a complex neural network (decoder or generator) to generate the final samples. Given that the neural network already provides a complex mapping, **choosing a simple distribution (such as a standard normal distribution) for the latent variables makes sampling convenient and efficient**.

However, this design also imposes limitations on the capabilities of VAE and GAN. Ideally, a generative model should uncover the **causal relationships** between the latent variables **z** and the target datasets $\mathbf{X}_{data}$. This would allow the model to infer meaningful, interpretable information from the latent space (expressed through the posterior distribution of the latent variables). However, using a simple prior distribution like a standard normal distribution **restricts the model's ability to infer the true latent distribution, its invertibility, and its causal interpretability**, because the true underlying latent distribution may be significantly different from the assumed normal distribution prior.

For VAE in particular, its theoretical foundation is **Bayesian inference** (Bishop and Nasrabadi (2006); Murphy (2012)), which aims to infer the underlying parameters or causes of observable data and phenomena. A purely neural network-driven approach that relies solely on **brute-force fitting** without interpretability is insufficient. However, in **vanilla VAE**, the prior distribution of the latent variables is fixed as a standard normal distribution (or a similarly simple distribution), and its parameters are neither trainable nor adaptive. As a result, the posterior distribution may **force to approximate a unreasonable prior**, leading to suboptimal performance. In particular, when the standard normal distribution is used as the latent prior, the posterior distribution is constrained in a way that can cause **blurry** generated samples (Tolstikhin et al. (2017)), as illustrated in Figure 5.

In Bayesian theory, the choice of prior is crucial for inference quality. **Setting the prior as a simple standard normal distribution limits the inference capacity of VAE**, which in turn affects its generative capabilities. In fact, **inference and generation are inherently linked**—better causal inference leads to more meaningful and realistic generated samples. This interplay between inference and generation will be explored further in subsequent articles.

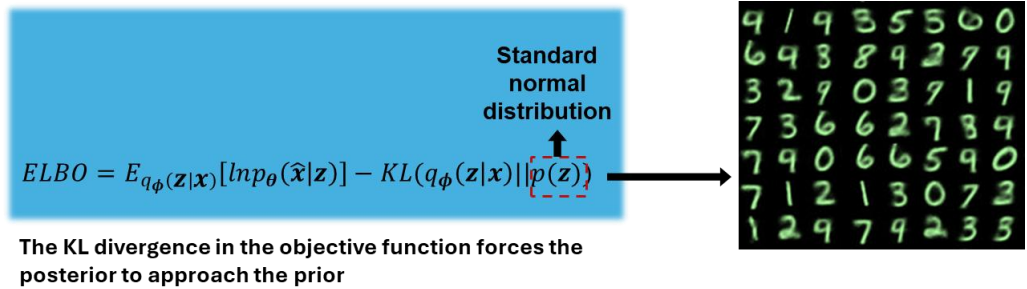

Figure 5. Unreasonable prior settings affect VAE's inference ability and generation ability: The posterior of the latent variable is forced to approximate the standard normal distribution prior with fixed parameters, causing VAE to lose its variational inference ability, and the KL divergence term becomes a **regularization term** (because the lack of VAE's inference ability affects its generative expression ability, making the image blurry).

## 5. Conclusion
1. **Fundamental Nature of Vanilla VAE and GAN**

    This paper explores the essence of vanilla VAE and GAN, which is based on using a simple

base distribution combined with a neural network mapping to approximate arbitrarily complex distributions. This approximation is implicit, as neither VAE nor GAN can explicitly express the shape of the target distribution.

2. **Necessity and Rationality of Implicit Approximation**
   The paper briefly explains why **implicit approximation is both necessary and reasonable** for VAE and GAN. Since high-dimensional data distributions (e.g., facial images) are often too complex to be explicitly formulated, **implicit modeling is a practical solution** that allows generative models to learn underlying structures without requiring an explicit probability density function.

3. **Advantages and Disadvantages of Using a Simple Basis Distribution**
   - **Advantages**: Choosing a simple distribution (e.g., a standard normal distribution) as the latent prior **facilitates sampling**, making the implementation and training of generative models more convenient.
   - **Disadvantages**: (i): The true latent distribution may differ significantly from a simple prior, which **limits the model's inference capability and interpretability**. (ii): Since inference and generation capabilities are inherently linked (analogous to a conjugate relationship), **a weaker inference ability can indirectly degrade the quality of generated samples**.

This paper provides a foundational understanding of how VAE and GAN perform implicit distribution approximation, the necessity of this approach, and the trade-offs of using simple base distributions in generative models.